\documentclass{article}

\usepackage[final]{corl_2021} 

\usepackage[hyperref]{}
\usepackage{times}
\usepackage{lipsum}
\usepackage{latexsym}
\usepackage{amsmath} 
\usepackage{amssymb}  
\usepackage{graphics} 
\usepackage{epsfig} 
\usepackage{mathptmx} 

\usepackage[utf8]{inputenc}
\usepackage{booktabs}
\usepackage{multirow}
\usepackage{subcaption} 
\usepackage{array}
\usepackage{xspace}
\usepackage{url}
\usepackage{soul}
\usepackage{xcolor}
\usepackage{pifont}
\usepackage{graphicx}
\usepackage{wrapfig}
\usepackage{enumitem}
\usepackage{colortbl}

\usepackage{microtype}

\newcommand{\blank}{$\rule{0.5cm}{0.15mm}$}
\newcommand{\textquoteinline}[1]{\textit{#1}}
\newcommand{\objclass}[1]{\texttt{#1}}

\newcommand{\dataset}{\textsc{SNARE}}
\newcommand{\datasetfull}{\textbf{S}hape\textbf{N}et \textbf{A}nnotated with \textbf{R}eferring \textbf{E}xpressions}
\newcommand{\modelfull}{\textbf{La}nguage \textbf{G}rounding through \textbf{O}bject \textbf{R}otation}
\newcommand{\model}{\textsc{LaGOR}}
\newcommand{\scorerfull}{Language-View Match}
\newcommand{\scorer}{\textsc{Match}}

\newcommand{\vis}{visual}
\newcommand{\nonvis}{blindfolded}
\newcommand{\Vis}{Visual}
\newcommand{\Nonvis}{Blindfolded}
\newcommand{\NV}{Blind}

\newcommand{\vilbert}{ViLBERT}

\definecolor{Gray}{gray}{0.90}
\newcolumntype{a}{>{\columncolor{Gray}}r}
\newcolumntype{g}{>{\columncolor{Gray}}c}
\newcommand{\B}[1]{\textcolor{blue}{\textbf{#1}}}

\title{Language Grounding with 3D Objects}

\author{
    Jesse Thomason\thanks{Equal contributions.} \\
    University of Southern California
   \And
  Mohit Shridhar\textnormal{$^*$} \\
  University of Washington
  \AND
  Yonatan Bisk \\
  Carnegie Mellon University
  \And
  Chris Paxton \\
  NVIDIA 
  \And
  Luke Zettlemoyer \\
  University of Washington
}

\date{}

\hypersetup{urlcolor=blue, colorlinks=true}

\begin{document}
\maketitle

\begin{abstract}

Seemingly simple natural language requests to a robot are generally underspecified, for example \textquoteinline{Can you bring me the wireless mouse?}
Flat images of candidate mice may not provide the discriminative information needed for \textquoteinline{wireless}.
The world, and objects in it, are not flat images but complex 3D shapes.  
If a human requests an object based on any of its basic properties, such as color, shape, or texture, robots should perform the necessary exploration to accomplish the task. 
In particular, while substantial effort and progress has been made on understanding explicitly visual attributes like color and category, comparatively little progress has been made on understanding language about shapes and contours.
In this work, we introduce a novel reasoning task that targets both visual and non-visual language about 3D objects.  
Our new benchmark \datasetfull{} (\dataset{}) requires a model to choose which of two objects is being referenced by a natural language description.\footnote{\url{https://github.com/snaredataset/snare}}
We introduce several CLIP-based~\cite{clip} models for distinguishing objects and demonstrate that while recent advances in jointly modeling vision and language are useful for robotic language understanding, it is still the case that these image-based models are weaker at understanding the 3D nature of objects -- properties which play a key role in manipulation.
We find that adding view estimation to language grounding models improves accuracy on both \dataset\ and when identifying objects referred to in language on a robot platform, but note that a large gap remains between these models and human performance.
\end{abstract}


\section{Introduction}
\begin{wrapfigure}[16]{r}{0.30\textwidth}
    \vspace{-15pt}
    \centering
    \includegraphics[width=\linewidth]{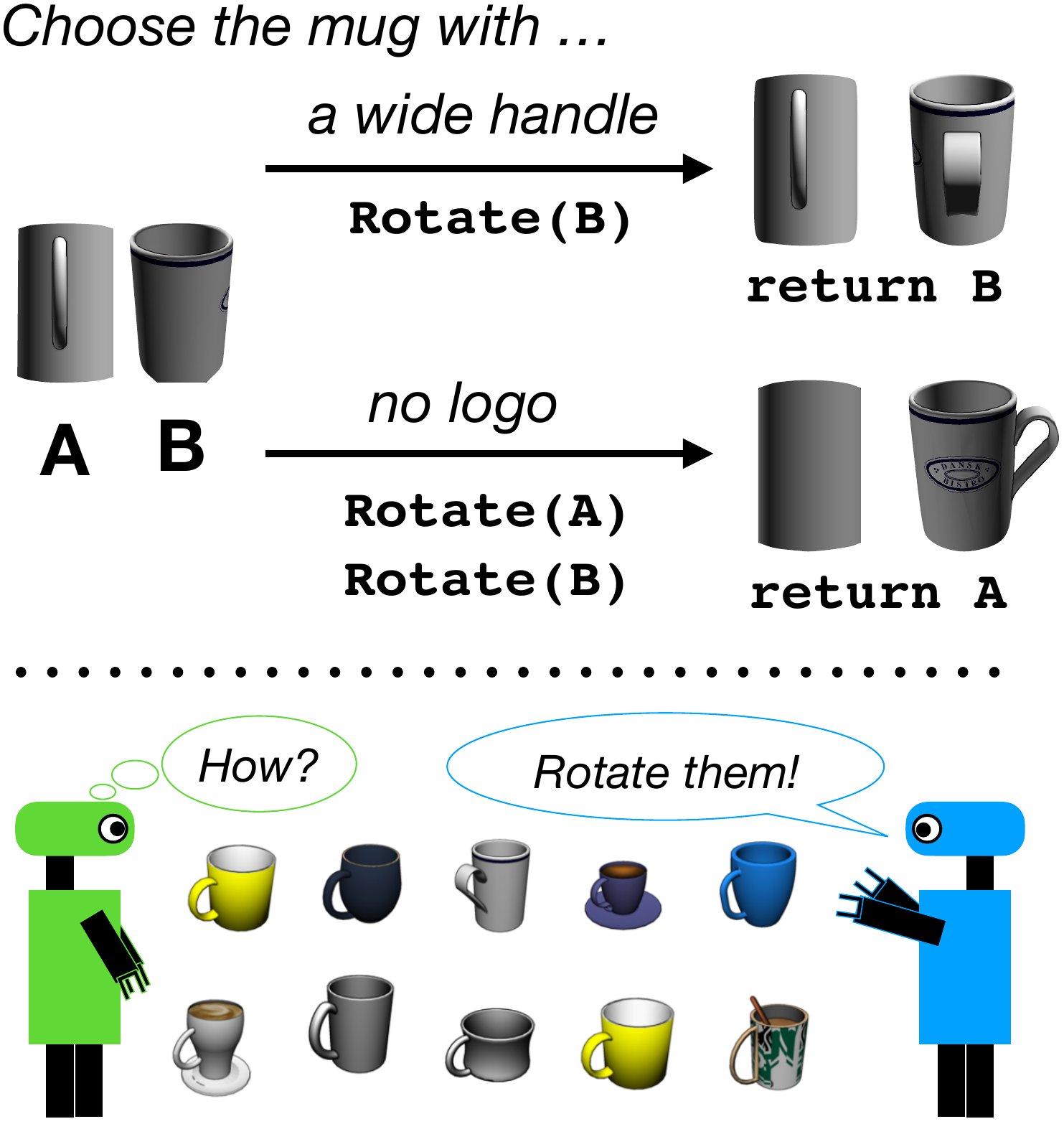}
    \caption{Deciding between 3D objects described in natural language involves rotating objects to see multiple faces.}
    \label{fig:main}
\end{wrapfigure}

Joint language and vision models are often trained on image captions which have a bias towards canonically ``visual" attributes of objects, such as color, rather than functional ones like shape.
Image captions omit properties that require understanding objects as 3D, rather than flat, concepts.  
In this work, we show how the effect of this disconnect is that while robotics research has benefited greatly from advances in computer vision, vision-and-language models cannot always be directly applied to robotics.
People use geometric and physical properties of objects when describing them.
For example, a robot carrying out a request to \textquoteinline{Bring me the mug with the wide handle}
needs to be able to spot the \textquoteinline{wide handle}, even if the mug rests on a countertop with the handle oriented out of view, but a robot can rotate such an object to investigate in further detail.
Generally, identifying objects based on natural language is more challenging from non-canonical viewpoints, which are common for robots in home, office, and industrial environments.

The codification and commodification of ResNet~\cite{resnet} architectures pretrained on ImageNet~\cite{imagenet} has yielded immense progress in off-the-shelf computer vision techniques.
Such representations provide a strong foundation for visual reasoning.
With the introduction of pretrained Transformer~\cite{vaswani2017attention} architectures like BERT~\cite{bert}, natural language processing has similar off-the-shelf tools for representing language.
Modern visual language grounding is done through massive, internet-scale pretraining to combine these strengths: aligning images with their captions and content descriptions in natural language using Transformer models~\cite{lu2019vilbert,clip,uniter,visualbert}.

However, internet images suffer from a type of \textit{reporting bias}.
These images are captured by sighted humans, from human-centric vantage points~\cite{torralba2011unbiased}.
There is a domain shift between such canonical depictions of objects in images and objects as seen through a robot camera~\cite{kanade2012first,vln-sim2real}, as studied in computer vision between such images and those captured by the blind~\cite{vizwiz}.
For example, images of \textquoteinline{mugs} online almost exclusively feature a full or three-fourths view of the handle, while a mug in the wild can easily be oriented such that the handle is occluded from the camera by the mug body.

A person requesting a mug with a wide handle has a \textit{mental model} of the referent object that includes its full 3D spatial information; humans do not imagine objects only in 2 dimensions.
Further, someone trying to retrieve said mug will differentiate it from other mugs on the countertop by viewing the mugs from different angles to inspect the handles.

Our goal is to similarly imbue robots with three dimensional notions of language grounding, by encouraging them to \textit{rotate} an object when selecting the referent of a language description.
We introduce the \datasetfull\ (\dataset) dataset, which provides discriminative natural language descriptions of 3D ShapeNet~\cite{shapenet2015} object models.
\dataset\ includes both \textit{\vis} descriptions that focus on colors and object categories and \textit{\nonvis} descriptions that focus on shapes and parts.
We finetune large off-the-shelf models to select the correct object given its description, and find that estimating the input views as an auxiliary prediction improves accuracy.

We introduce \modelfull\ (\model) which performs view estimation as an auxiliary loss to encourage 3D object understanding (Section~\ref{sec:methods}) during the \dataset\ task.
We find that using multiple views to select the correct object improves over single-view object selection, using the large-scale vision-and-language CLIP~\cite{clip} model as a backbone to score how well an image and language \scorer.
We show that models taking in multiple views of objects objects, while performing auxiliary view estimation, can result in accuracy approaching that of models consuming panoramic views of objects, while also being more realistic for a robot.

Our key contributions are:
\setlength{\leftmargini}{0.5cm}
\begin{itemize}[noitemsep,nosep]
    \item \dataset, a benchmark dataset for identifying 3D object models given natural language referring expressions including \textit{visual} and \textit{blindfolded} descriptions; 
    \item Baseline models for \dataset\ demonstrating both zero-shot and fine-tuned performance of state-of-the-art vision-and-language models; and
    \item \model, an initial \dataset\ model that looks at two random views of an object, estimates the angle of those views, then makes a referent prediction, which we demonstrate using a robot.
\end{itemize}

\section{Related Work}

Using natural language to work with robot partners is a long-standing goal in robotics~\cite{tellex:arcras:20}.
We argue that internet-scale, pretrained vision-and-language models offer a powerful starting point for human-robot collaboration.
Unlike 2D internet images, physical objects can be picked up and moved by robot agents.
Agents can perform information-seeking behaviors on objects to improve their world~\cite{ogata-motor-babbling,mousavian20196,interaction-exploration} and language understanding~\cite{hill2020human,thomason:jair20,lynch:arxiv20}.
Such world interaction is inextricable from language grounding~\cite{bisk2020experience}, motivating language annotations for higher-fidelity referents than static images.
We introduce \dataset, comprised of language referring expressions for 3D objects, and the \model\ model to combine language and multiple object views to achieve better 3D understanding.
\dataset\ extends lines of work tying language to static images, 3D object models, and even physical objects.
We demonstrate that \model\ generalizes to physical objects manipulated by a tabletop robot \S\ref{ssec:robot_results}.
\model\ is inspired by models that perform information-seeking actions informed by learned world models to better achieve goals.

\textbf{Image-based Object Identification} 
Object classification~\cite{imagenet,resnet} is the first step towards image captioning and visual question answering~\cite{bottom-up-top-down}.
Particular object instances can be found with region segmentation models such as MaskRCNN~\cite{maskrcnn}, enabling object referent tasks such as GuessWhat!?~\cite{guesswhat}.
Combining visual recognition with Transformer-based language understanding to jointly attend to language and visual tokens leads to improved downstream performance on many vision-and-language tasks~\cite{lu2019vilbert,Lu_2020_CVPR,visualbert}.
Joint embedding approaches that learn a shared subspace for representing language and vision tokens~\cite{leong-mihalcea-2011-going} also achieve state-of-the-art performance when trained at scale~\cite{clip}.
We show that these large-scale, pretrained models fall short of human performance on the \dataset\ task, and that synthesizing 2D views from multiple vantage points improves object identification performance.

\textbf{Language Grounding in 3D}
Prior works have associated single-word attributes with 3D object models based on latent representations of 3D meshes~\cite{cohen:iros19:grounding_eigenobjects}.
To learn spatial language, vision-and-language navigation (VLN)~\cite{vln} models infer navigation actions from instructions and visual observations in 3D simulated worlds.
Such tasks can be extended to include simulated world interactions such as picking and placing objects~\cite{imitating_interactive_intelligence} and using appliances and tools~\cite{alfred}.
Models for these tasks extend Transformer representations to include action taking~\cite{majumdar:eccv20,cyclevlnbert} and invoke object classification methods to create a semantic understanding of the world~\cite{moca}.
\dataset\ poses a complementary challenge, providing data for selecting referent objects in the presence of distractors by taking into account the multiple views possible of objects in 3D space.

\begin{wraptable}[13]{r}{0.55\linewidth}
\vspace{-12pt}
\centering
\begin{tabular}{clrrr}
    \textbf{Data} & \textbf{Fold} & \textbf{\# Cats} & \textbf{\# Objs} & \textbf{\# Ref Exps} \\
    \toprule
    \multirow{3}{*}{\footnotesize \rotatebox{90}{ShapeGlot}} & Chairs & 1 & 4,511 & 78,789 \\
    & Other & 4 & 200 & 400 \\
    \cmidrule{2-5} 
    & \textbf{Total} & 5 & 4,711 & 79,189 \\
    \midrule
    \multirow{5}{*}{\rotatebox{90}{\textbf{\dataset}}} & Train & 207 & 6,153 & 39,104 \\
    & Val & 7 & 371 & 2,304 \\
    & Test & 48 & 1,357 & 8,751 \\
    \cmidrule{2-5}
    & \textbf{Total} & 262 & 7,881 & 50,159 \\
    \bottomrule
\end{tabular}
\caption{
    Fold summaries in \dataset, with the ShapeGlot benchmark size for contrast.
}
\label{tab:fold_stats}
\end{wraptable}

The work most similar to ours is ShapeGlot~\cite{shapeglot}, a collection of referring expressions for ShapeNet objects to discriminate between two distractors.
While ShapeGlot is similar in spirit, it contains training data only for \objclass{chair} objects, and tests on four additional categories.
By contrast, \dataset\ spans 262 distinct object categories (Table~\ref{tab:fold_stats}).
ShapeGlot focuses on a model's abilities to learn particular parts of objects, such as chair arms and legs, while \dataset\ aims to ground language more generally to 3D features of objects spanning color, shape, category, and parts.
Further, ShapeGlot models take in a pointcloud representation of objects along with visual features, where our \model\ model does not assume access to the 3D model of an object and instead operates solely on different 2D views achieved through object rotation.

\textbf{Physical Object Identification}
Modeling the connections between language to physical objects enables robots to identify objects for manipulation by color, shape, and category~\cite{ingress,Kollar2017GeneralizedGG}.
The physical forces and sounds objects make during manipulation actions can also be associated with words such as \textquoteinline{rattling} and \textquoteinline{heavy} for multimodal understanding beyond vision~\cite{thomason:jair20,thomason:ijcai16}.
Prior work has gathered language annotations for the YCB Benchmark object set~\cite{ycb} to explore how language descriptions provide priors on object affordances~\cite{scalise2019improving}.
In our tabletop robot experiments, we use camera views of novel objects to evaluate zero shot transfer of \model\ to robot camera views of real objects, which are processed by an off-the-shelf segmentation algorithm.

\textbf{Information-Seeking Actions}
Embodiment, whether in simulation or the physical world, affords agents the opportunity to seek out new information to help with a given task.
By predicting the new information that can be gotten from different actions, agents can \textit{prospect} over potential futures for planning~\cite{planning-multimodal-sinapov,pmlr-v119-nair20a,paxton2019prospection}.
In the aforementioned VLN task, predicting \textit{what} will be seen along different potential routes facilitates more efficient navigation~\cite{pathdreamer}.
Our \model\ model estimates the angles of object input views, encouraging 3D understanding in the learned object representations.

\section{\datasetfull\ (\dataset)}
\label{sec:dataset}

\begin{figure}[t]
\centering
\includegraphics[width=\textwidth]{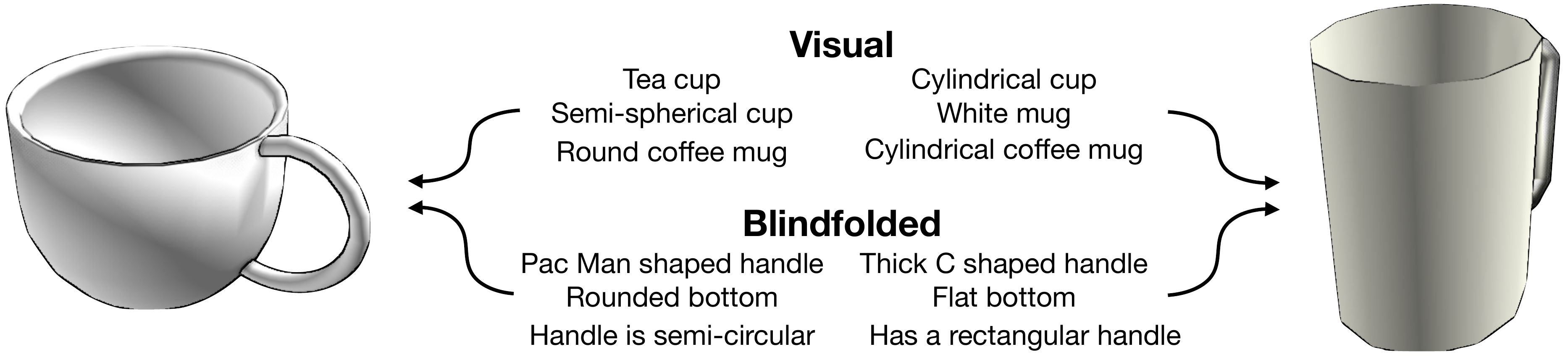}
\caption{
    Example object pair and object referring expressions in \dataset.
    Given a referring expression and one (or more) views of two contrasting objects, a model must decide which object is being referenced by the language. Language annotations are collected in two forms: \textbf{Visual} and \textbf{Blindfolded}.  The latter expressions tasked annotators with describing objects tactilely.
}
\vspace{-10pt}
\label{fig:object_and_re_samples}
\end{figure}

We introduce \dataset{}, as a new benchmark for grounding natural language referring expressions to distinguish 3D objects.  
The annotations are collected to complement the ACRONYM\footnote{Yes, that dataset title is ``ACRONYM''~\cite{acronym2020}.} \cite{acronym2020} grasping dataset and include language that targets both visual and tactile attributes of objects.
Our goal is to enable future research to ground both from multi-view vision, as in this work, and directly from robotic grasp contact point data (Section~\ref{sec:conclusions}).

To construct \dataset, we select a subset of $7,897$ ACRONYM \cite{acronym2020} object models from ShapeNetSem~\cite{chang:fpic15,shapenet2015}.
We obtain over 50,000 natural language referring expressions in English for these object models via Amazon Mechanical Turk (AMT).\footnote{Section~\ref{ssec:amt} contains additional details about the AMT study.}
Figure~\ref{fig:object_and_re_samples} gives an example of referring expressions collected for two \objclass{mug} objects used to distinguish between the two.

To elicit referring expressions with high specificity, we frame the annotation as a discriminative task.
AMT workers were presented with two object models side-by-side from the same ShapeNet category, for example two different \objclass{OutdoorTable} object meshes, and asked to complete sentences like
\begin{itemize}[noitemsep,nosep]
 \item The way to tell Object A from Object B is that Object A looks like a(n) \blank
 \item Blindfolded, the way to tell Object A from Object B is that Object B is a(n) \blank
\end{itemize}
with referring expressions.
We displayed the object models as GIFs rendered to give a 360 rotating view of each object.
The resulting \dataset\ task asks models to take in one such referring expression and decide whether it applies to Object A or Object B.

Priming in the annotation prompts invokes visual features (\textquoteinline{\dots looks like\dots}) and non-visual features (\textquoteinline{Blindfolded, \dots}) in the referring expressions.
By pairing objects from the same ShapeNet category, referring expressions must go beyond categorical information like \textquoteinline{brown dresser} to differentiate one object from the other with higher specificity, similar to ShapeGlot's choice to use distractor objects from a single training category~\cite{shapeglot}.
We collected six referring expressions per object---three primed to be \vis\ and three primed to be \nonvis.
Priming for \vis\ expressions mentioned \textquoteinline{the name of the object, shapes, and colors}, while for \nonvis\ mentioned \textquoteinline{shapes and parts}.
Each referring expression was vetted through a secondary task on AMT where, given the language expression, workers had to correctly select the referent object.
Every referring expression in \dataset\ was correctly associated to its referent object by a majority of such annotators.

\dataset\ referring expressions average $4.27$ words, with \nonvis\ expressions longer ($4.95$) than \vis\ expressions ($3.63$).
We estimated that \nonvis\ expressions use more shape words (14\% vs 5\% of all words) while \vis\ use more color words (22\% vs 1\% of all words), by traversing the WordNet~\cite{wordnet} hierarchy for each word and noting whether it is a hyponym of \textquoteinline{color} or \textquoteinline{shape}.
This distribution difference is consistent with the priming used to elicit \vis\ and \nonvis\ expressions.

Each \dataset\ instance is a tuple of (referring expression, Object A, Object B), and models must select which of Object A or B the referent of the natural language expression.
We split these data instances into train, validation, and test folds by ShapeNet category.
We ensure that closely related categories such as \objclass{2Shelves} and \objclass{3Shelves} or \objclass{DiningTable} and \objclass{AccentTable} are within the same fold.
For example, all shelf-related and bed-related categories are sorted into the train and test folds, respectively, so that inter-category information does not leak across folds.
More details can be found Section~\ref{ssec:data_folds}.
Table~\ref{tab:fold_stats} summarizes the overall data statistics.

\section{Methods}
\label{sec:methods}

\begin{figure}[t]
\centering
\begin{small}
\begin{tabular}{cccccc}
    \raisebox{-.5\height}{\includegraphics[width=0.16\textwidth]{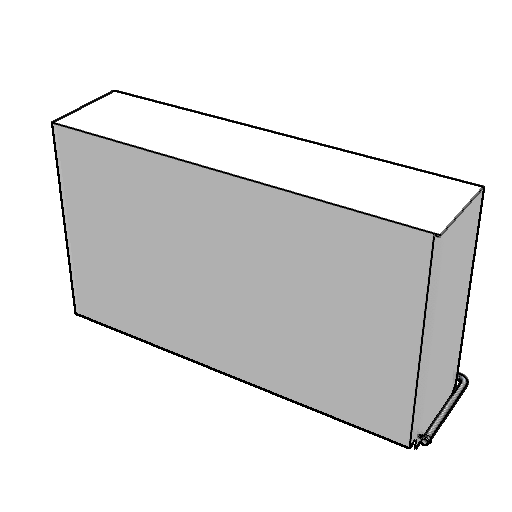}} &
    $\overrightarrow{\text{90}^{\circ}\circlearrowright}$ &
    \raisebox{-.5\height}{\includegraphics[width=0.16\textwidth]{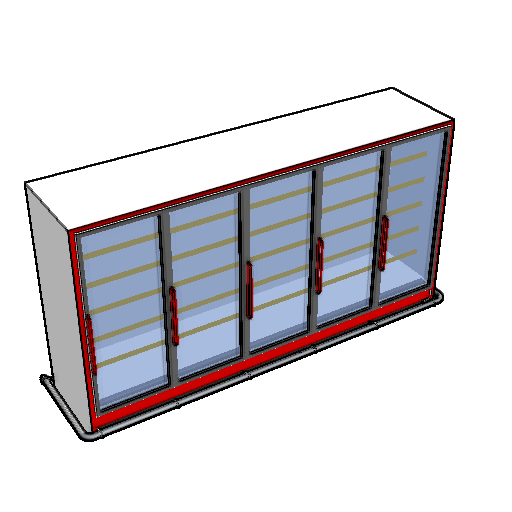}} &
    \raisebox{-.5\height}{\includegraphics[width=0.16\textwidth]{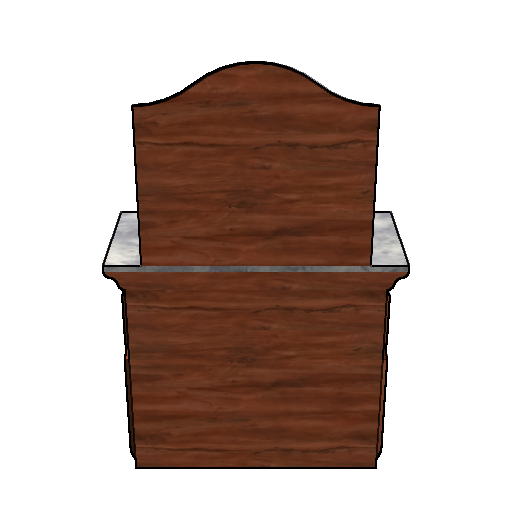}} &
    $\overrightarrow{\text{180}^{\circ}\circlearrowright}$ &
    \raisebox{-.5\height}{\includegraphics[width=0.16\textwidth]{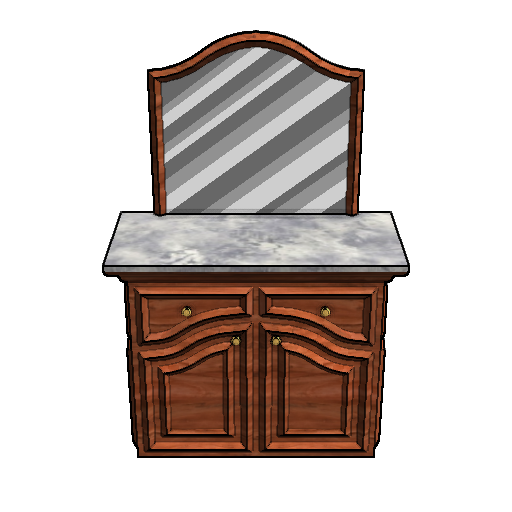}} \\
    \multicolumn{3}{c}{\textbf{\Vis} - \textquoteinline{red gray and white display case}} & 
    \multicolumn{3}{c}{\textbf{\NV} - \textquoteinline{cube bottom with 4 knobs}} \\
    \multicolumn{3}{c}{\scorer\ score after rotation $+1388\%$} &
    \multicolumn{3}{c}{\scorer\ score after rotation $+246\%$} \\
    \raisebox{-.5\height}{\includegraphics[width=0.16\textwidth]{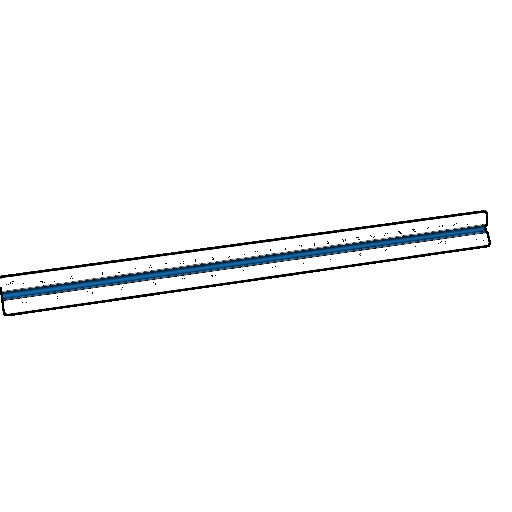}} &
    $\overrightarrow{\text{135}^{\circ}\circlearrowright}$ &
    \raisebox{-.5\height}{\includegraphics[width=0.16\textwidth]{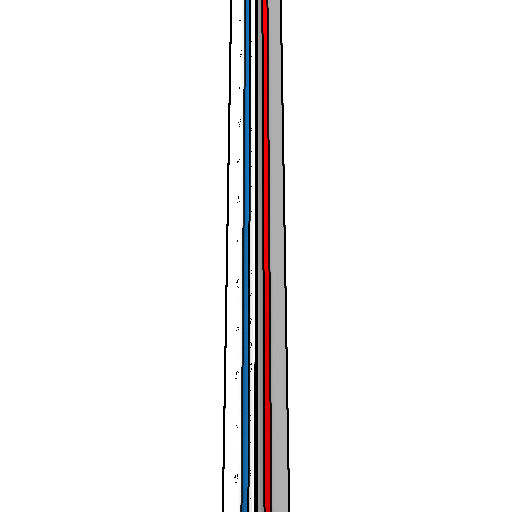}} &
    \raisebox{-.5\height}{\includegraphics[width=0.16\textwidth]{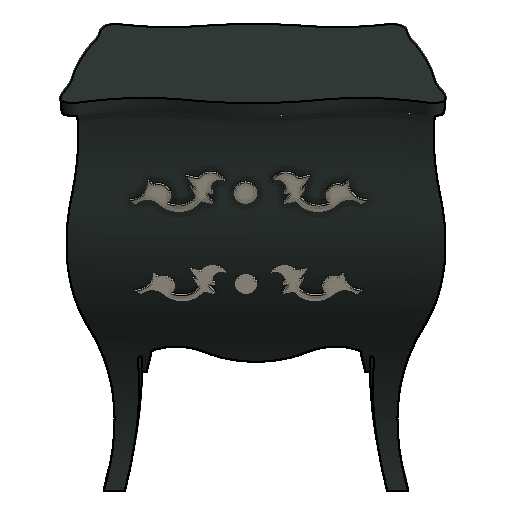}} &
    $\overrightarrow{\text{45}^{\circ}\circlearrowleft}$ &
    \raisebox{-.5\height}{\includegraphics[width=0.16\textwidth]{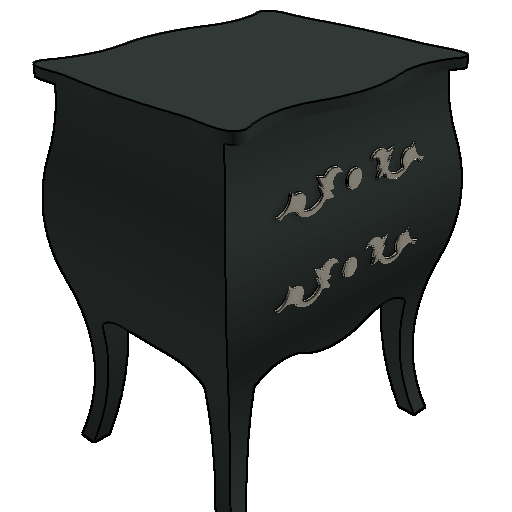}} \\
    \multicolumn{3}{c}{\textbf{\Vis} - \textquoteinline{red white and blue device}} & 
    \multicolumn{3}{c}{\textbf{\NV} - \textquoteinline{nightstand with smooth flat top and four legs}} \\
    \multicolumn{3}{c}{\scorer\ score after rotation $+131\%$} &
    \multicolumn{3}{c}{\scorer\ score after rotation $+205\%$} \\
\end{tabular}
\end{small}
\caption{
    Example \scorer\ score increases after performing object rotation.
    Rotations involve exposing colors (top left, bottom left), canonical faces (top right), and parts (bottom right).
    A robot encountering objects in the wild will find initial object orientations that do not line up with natural language referring expressions for those objects, as reflected in \dataset.
}
\label{fig:object_rotations}
\end{figure}

There has been a recent proliferation of off-the-shelf, increasingly large-scale pretrained vision and language alignment models applicable for language grounding in robotics~\cite{clip,lu2019vilbert,uniter,visualbert,zhang2021vinvl}.
We set out to answer two questions about such models, using \dataset\ as a testbed:
\begin{itemize}[noitemsep,nosep]
    \item Can existing language and vision models ground language in the \dataset\ benchmark?
    \item Do \dataset\ models generalize to a robot object selection task better than off-the-shelf models?
\end{itemize}
To answer the first question, we first train a \scorer\ module that learns a predictive head on top of a CLIP~\cite{clip} backbone, and evaluate this module on single and multiple views of 3D objects.
To answer the second question, we introduce the \model\ model, which uses view estimation as an auxiliary loss while predicting language expression referent objects for \dataset, and evaluate on both \dataset\ \S\ref{ssec:snare_results} and a physical robot platform \S\ref{ssec:robot_results}.
\model\ examines an initial 3D object view and an additional, post-rotation view, striking a balance between single-view models and those that attempt to capture each object's entire 3D structure.

\subsection{\scorerfull\ (\scorer) Module} \label{sec:classifier}

The \scorer\ module takes in a language expression $\sc{L}$ and single view of an object, $\sc{V}_i$, and produces a match score $s(\sc{L}, \sc{V}_i)$.
To decide whether Object A or Object B is the referent of a language expression on \dataset, we interpret  $\text{argmax}_{O\in\{A,B\}}s(\sc{L}, \sc{V}_{i,O})$ as the model's prediction.

CLIP~\cite{clip} serves as a backbone on which we add additional, learnable layers for the \dataset\ task.
We use CLIP's transformer-based sentence encoder to extract language features $\mathbf{l}~:~\mathbb{R}^{1 \times 512}$.
We use CLIP ViT-B/32 to extract visual features for image $\sc{V}_{i,A}$, $\mathbf{v}_{\text{i,A}}~:~\mathbb{R}^{1 \times 512}$, and image $\sc{V}_{i,B}$, $\mathbf{v}_{\text{i,B}}~:~\mathbb{R}^{1 \times 512}$. 
These modality-specific feature vectors are concatenated: $[\mathbf{v}_{\text{i,A}};\mathbf{l}]~:~\mathbb{R}^{1 \times 1024}$ and $[\mathbf{v}_{\text{i,B}};\mathbf{l}]~:~\mathbb{R}^{1 \times 1024}$ and independently run through a learnable, multi-layer perceptron that gradually reduces the dimensionality from 1024 to 512, then 256, and finally to a single-dimensional score $s$. This final score comparison training mirrors the multiple-choice formulation used in existing unimodal \cite{bert} and multimodal transformers \cite{uniter}.
We keep the CLIP encoders frozen during training.

ShapeNet objects are 3D meshes, incompatible with the 2D input expected by off-the-shelf vision and language models.
We sample eight rendered viewpoints around each object at 45 degree increments, and compare off-the-shelf CLIP and \scorer\ module accuracies when considering single or multiple of these views.
To aggregate multiple views, we \texttt{maxpool} over view embedding vectors $\mathbf{v}_{\text{i,A}}\dots\mathbf{v}_{\text{j,A}}$.

\scorer\ is trained with cross-entropy-loss, $\mathcal{L}_s$, to predict a binary label of whether the referring expression matches the object represented in the view image.
\scorer\ is trained for 50 epochs on \dataset, with object views chosen at random during each step. 

\textbf{Baselines.} We compare our fine-tuned \scorer\ against a zero-shot CLIP classifier. 
Instead of fine-tuning CLIP, we use the cosine distance between visual and language features to pick the referred object. 
That is, we select $\text{argmax}_{O\in\{A,B\}}\mathbf{l}\cdot\mathbf{v}_{\text{i,O}}$.
We also evaluate against a trained \vilbert\ baseline that consumes multiple object views at once, but find that it does not perform as well as the \scorer\ module; more details about the \vilbert\ baseline can be found in Section~\ref{ssec:vilbert}.
The \scorer\ model can be thought of as a fine-tuned CLIP model; rather than back-propagating through the CLIP representation itself we learn additional tuning layers on top.

\subsection{\modelfull\ (\model)}

\begin{figure*}[!t]
    \centering
    \vspace{-15pt}
    \includegraphics[width=.9\linewidth]{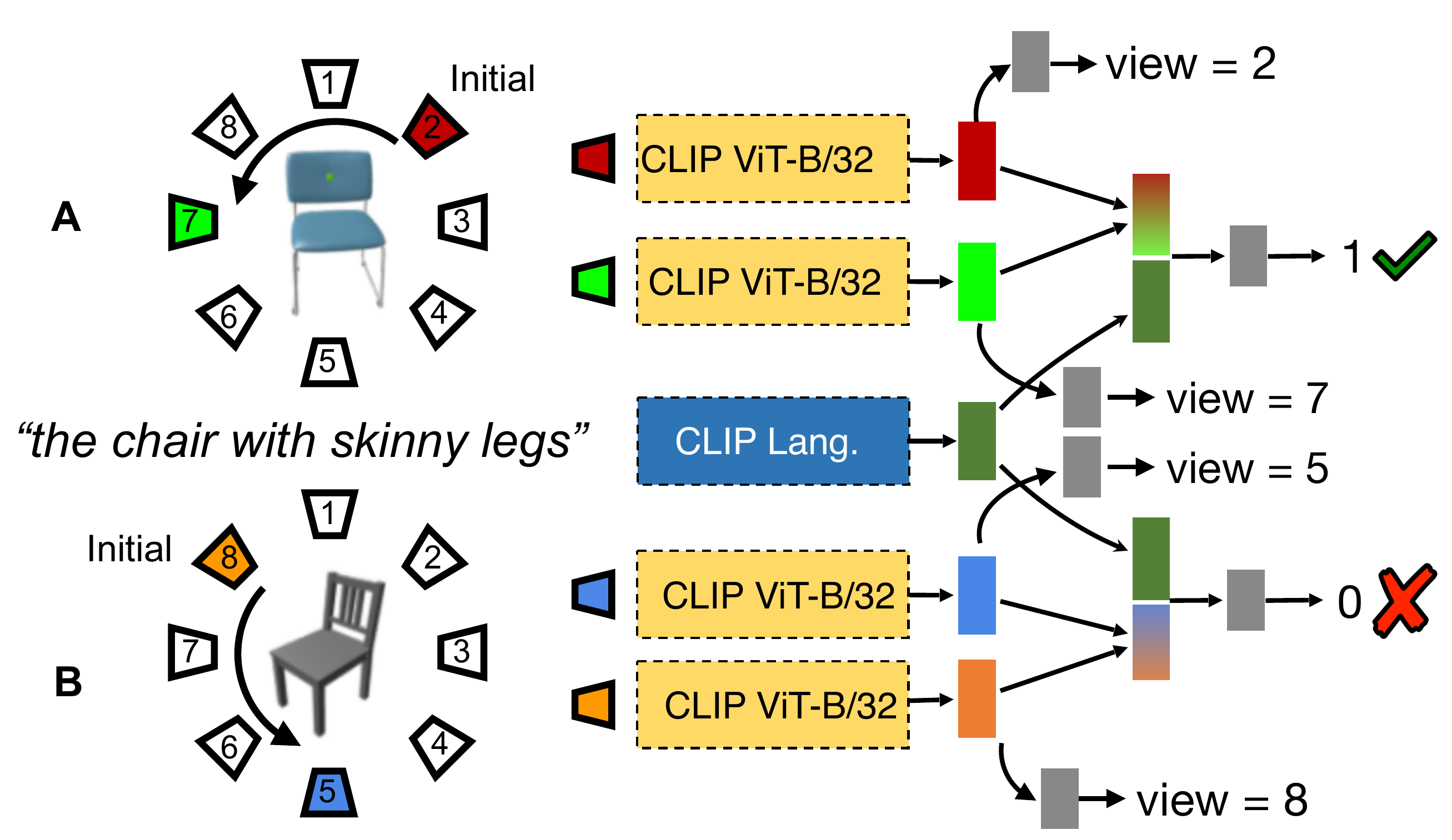}
    \caption{\textbf{\model\ model} relies on the pretrained CLIP~\cite{clip} architecture as a backbone encoder across multiple views compared to the encoded referring expression. 
    The model sees improvements from auxiliary losses that predict the orientation of the initial and final view.}
    \label{fig:model}
\end{figure*}

A robot tasked with retrieving an object given a natural language expression should not have to gather a 360 degree view of each candidate referent object to make a decision.
We have the intuition that estimating the input viewing angles can enable a model to develop a global reference frame for language grounding against 3D objects (Figure~\ref{fig:object_rotations}).
Thus, we propose \model, which takes in two views of each candidate object, performs \textit{view estimation} on those views as an auxiliary prediction (Figure~\ref{fig:model}), then predicts the language referent.

In addition to predicting the object referent using a pre-trained \scorer\ module that considers two views, \model\ predicts the input views of each object using a cross-entropy loss, $\mathcal{L}_v$, against a 1-hot vector representing a discrete set of 8 views at 45 degree offsets.
\model\ learns a multi-layer perceptron that takes in a visual embedding $\sc{V}_{i}$ and reduces dimensionality from 512 to 256 to 128 to 64 to a vector of 8 logits representing the 8 discrete object views.
The \scorer\ loss, $\mathcal{L}_s$, is combined with the view estimation loss for a final loss function $\mathcal{L}=\mathcal{L}_v + 0.2*\mathcal{L}_s$.

\subsection{Robot Demonstration}
We evaluate \model\ trained on \dataset\ on a robot platform, taking two random views of objects and evaluating the robot's ability to select which is the correct referent of a language expression (Section~\ref{ssec:robot_results}).
We compare \model\ to the off-the-shelf CLIP model that considers a single view of the object only, before performing any view-gathering rotations.

We set two objects on the workspace of a Franka Emika Panda with a wrist-mounted Intel Realsense D414.\footnote{https://www.franka.de/ and https://www.intelrealsense.com/depth-camera-d415/ respectively}
We capture an initial view of the scene, then segment the objects from the workspace using the Unseen Object Clustering~\cite{xiang2020learning} algorithm.
We feed the segmented objects as the initial views to \model\ and CLIP.
We move the arm to a second vantage point above the workspace and capture another image of the objects and repeat segmentation to obtain the second views for \model.

\section{Results}

In this section we show that CLIP~\cite{clip} and \vilbert~\cite{lu2019vilbert} select correct 3D object referents in \dataset{} substantially less often than the \scorer\ and \model\ models.
Then, we deploy CLIP and \model\ on a robot tasked with selecting objects conditioned on natural language referring expressions, and find that view estimation and gathering additional object viewpoints improves referent identification.

\subsection{Performance on \dataset}
\label{ssec:snare_results}

\begin{table}
   \setlength{\aboverulesep}{0pt}
   \setlength{\belowrulesep}{0pt}
   \centering
   \begin{small}
    \begin{tabular}{l@{}caaarrr}
                    & & \multicolumn{3}{c}{\textbf{Validation}} & \multicolumn{3}{c}{\textbf{Test}}  \\
                    Model & Views & \multicolumn{1}{g}{\Vis$\subset$} & \multicolumn{1}{g}{\NV$\subset$} & \multicolumn{1}{g}{All} &
                    \multicolumn{1}{c}{\Vis$\subset$} & \multicolumn{1}{c}{\NV$\subset$} & \multicolumn{1}{c}{All} \\
    \toprule
    \vilbert & All & \bf 89.5\phantom{  $\pm 0.0$} & \B{76.6}\phantom{  $\pm 0.0$} & \B{83.1}\phantom{  $\pm 0.0$} & 80.2\phantom{  $\pm 0.0$} & \B{73.0}\phantom{  $\pm 0.0$} & \bf 76.6\phantom{  $\pm 0.0$} \\
    CLIP & All & 83.7  $\pm 0.0$ & 65.2  $\pm 0.0$ & 74.5  $\pm 0.0$ & 80.0  $\pm 0.0$ & 61.4  $\pm 0.0$ & 70.9  $\pm 0.0$ \\
    \scorer & All & 89.2  $\pm 0.9$ & 75.2  $\pm 0.7$ & 82.2  $\pm 0.4$ & \textbf{83.9}  $\pm 0.5$ & 68.7  $\pm 0.9$ & 76.5  $\pm 0.5$ \\
    \midrule
    CLIP & Single & 79.0  $\pm 0.0$ & 63.0  $\pm 0.0$ & 71.1  $\pm 0.0$ & 74.0  $\pm 0.0$ & 59.7  $\pm 0.0$ & 67.0  $\pm 0.0$ \\
    \scorer & Single & \textbf{88.4}  $\pm 0.4$ & \textbf{73.3}  $\pm 0.6$ & \textbf{80.9}  $\pm 0.4$ & \textbf{83.2}  $\pm 0.3$ & \textbf{68.0}  $\pm 0.5$ & \textbf{75.8}  $\pm 0.3$ \\
    \midrule
    CLIP & Two & 81.0  $\pm 0.0$ & 64.1  $\pm 0.0$ & 72.6  $\pm 0.0$ & 76.0  $\pm 0.0$ & 60.8  $\pm 0.0$ & 68.6  $\pm 0.0$ \\
    \scorer & Two & 89.2  $\pm 0.6$ & 74.4  $\pm 0.7$ & 81.8  $\pm 0.4$ & 83.7  $\pm 0.4$ & 68.7  $\pm 0.5$ & 76.4  $\pm 0.4$ \\
    \model & Two & \B{89.8}  $\pm 0.4$ & \textbf{75.3}  $\pm 0.7$ & \textbf{82.6}  $\pm 0.4$ & \B{84.3}  $\pm 0.4$ & \textbf{69.4}  $\pm 0.5$ & \B{77.0}  $\pm 0.5$ \\
    \midrule
    Human  (U) & All & 94.0\phantom{  $\pm 0.0$} & 90.6\phantom{  $\pm 0.0$} & 92.3\phantom{  $\pm 0.0$} & 93.4\phantom{  $\pm 0.0$} & 88.9\phantom{  $\pm 0.0$} & 91.2\phantom{  $\pm 0.0$} \\
    Human  (M) & All & 100.0\phantom{  $\pm 0.0$} & 100.0\phantom{  $\pm 0.0$} & 100.0\phantom{  $\pm 0.0$} & 100.0\phantom{  $\pm 0.0$} & 100.0\phantom{  $\pm 0.0$} & 100.0\phantom{  $\pm 0.0$} \\
    \bottomrule \\
    \end{tabular}
    \end{small}
    \caption{\textbf{Accuracy on the \dataset\ benchmark.}
    Mean accuracy $\pm$ standard deviation over 10 seeds.
    CLIP zero-shot accuracy is well above random chance, and adding more object views improves all models' performance.
    \model, which needs only two views, making it reasonable to deploy on a robot platform, outperforms two-view \scorer\ statistically significantly on both the validation and training set, by adding a level of 3D object understanding in the form of view estimation auxiliary losses.
    Human accuracy is conservatively calculated as the number of \dataset\ instances for which the correct referent object was identified by \textit{every} voting annotator (\textbf{U}nanimous); for all \dataset\ instances a majority of voting annotators correctly selected the referent (\textbf{M}ajority).
    }
    \label{tab:snare_results}
\end{table}

We train all models for 50 epochs on \dataset, with object views chosen at random during each step, and report the best validation performance across those epochs, as well as the test performance at that best validation epoch checkpoint.
We train each model 10 times with a different random seed to estimate performance variance across training runs.

Table~\ref{tab:snare_results} gives \model\ accuracy compared to other two-view alternatives, as well as models taking in all views or single views.
We compare to existing models trained (ViLBERT) and zero-shot (CLIP), as recommended in their respective papers.
We do not re-run ViLBERT training because it is extremely time-consuming; the same reason we build on CLIP for our model development.
To compare models, we perform an unpaired, Welch's two-tailed $t$-test on overall accuracy considering both \vis\ and \nonvis\ subsets.
We perform five such tests in all: for validation and test, \model\ against two-view \scorer, \model\ against all-view \scorer, and one additional pooling test (Section~\ref{ssec:pooling}).
We perform a conservative Bonferroni multiple-comparison correction applied to threshold of $p<0.05$.

\model\ statistically significantly outperforms two-view \scorer\ on both the validation ($p=0.0008$) and test ($p=0.0055$) sets.
The modelling difference between \model\ and \scorer\ with two views only the view estimation auxiliary losses.
\model\ averages higher accuracy than the \scorer\ that uses all views on every metric; however, these differences are significant for neither the validation ($p=0.0844$) or test ($p=0.0315$) set.\footnote{Note that with the Bonferroni correction, significance is reached only with $p<0.01$ given the desired threshold of $0.05$ with 5 tests run.} 

While no models achieve human level accuracy, the performance difference is particularly striking on the \nonvis\ referring expression subset of data, which lags 5-15\% behind \vis\ referring expressions across models.
That gap supports our intuition that \nonvis\ referring expressions capture a complementary and challenging linguistic space currently understudied for vision-language models but key to everyday manipulation. 
Thereby, \dataset{} opens interesting avenues for future work exploring the shape and grasp-points of objects.

Notably, zero-shot CLIP performance on \dataset, even when considering only one random view of each object, is well above random chance. 
While \scorer\ fine-tuning and \model\ auxiliary losses improve accuracy over zero-shot CLIP, the zero-shot performance is a good indicator that massively pre-trained language and vision alignment like CLIP could be useful for robotics tasks for which there is little or no in-domain data.

\subsection{Robot Results}
\label{ssec:robot_results}

\begin{figure}[t]
    \centering
    \includegraphics[width=1.\textwidth]{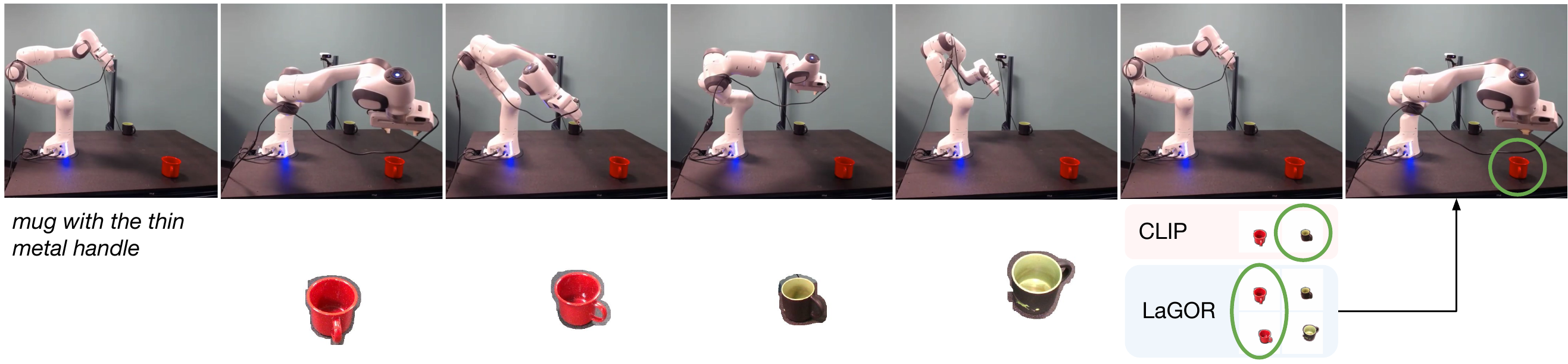}
    \caption{With a physical robot platform, \model\ outperforms zero-shot CLIP for selecting tabletop referent objects whose views are captured via multiple camera angles and off-the-shelf segmentation.}
    \label{fig:lagor_robot}
\end{figure}

We performed a set of experiments on a real robot (Figure~\ref{fig:lagor_robot}).
We run single view, zero-shot CLIP against \model\ on 11 referring expressions and pairs of objects.
Two objects at a time were placed on the table, and we selected two arbitrary viewing angles for each object.
We used unknown object instance segmentation~\cite{xiang2020learning} to segment out each object.
Zeroshot CLIP correctly selects the referent object 7 out of 11 times, while \model\ selects the correct referent 9 out of 11 times.
\model\ corrects CLIP's mistakes on expressions like \textquoteinline{Mug with the thin metal handle} where multiple views expose potentially occluded parts like \textquoteinline{handle}, as well as when information gathering makes decisions more confident, such as \textquoteinline{Orange box} when both objects have orange components, but one object is orange from all views.
These experiments introduce a number of new sources of noise not present in the simulated setting, for example changes in camera angle, shifting lighting during rotation, and automatic segmentation to account for cluttered scenes.
For example, segmentation leads to the creation of voids inside of objects not present in \dataset.
View extraction details, referring expressions, and extracted views are shown in Section~\ref{ssec:additional_robot}.

\section{Conclusions}
\label{sec:conclusions}

We introduce \datasetfull\ (\dataset), a challenge task to ground language to 3D object models.
We show that fine-tuned, massively pretrained vision and language models fall short of human performance at identifying object referents of natural language expressions by a wide margin, while still achieving well above random chance performance (Table~\ref{tab:snare_results}).
Models improve as more 2D views of 3D objects are available, but a robot can only capture one view at a time.
We introduce \modelfull\ (\model), a model that considers two views of candidate referent objects trained with current view estimation as an auxiliary loss, and builds on a pre-trained CLIP model.
We transfer the trained \model\ model to a physical robot to study its generalization to noisy images of physical objects.

In the future, a rotation \textit{policy} could be learned to examine up to $N$ views across $M$ candidate objects subject to a referring expression, rather than performing a random rotation to a novel view.
By assigning rotation actions expected discriminative values using a trained rotation module (though that could introduce a cyclic dependency with the \scorer\ module), and penalties based on the time it takes a particular hardware to achieve an object rotation, one could learn a POMDP policy aggregating object views seen so far to decide whether to next obtain a new view or make a guess about the referent object.
Obtaining different object views could be done either by picking up and rotating the object or by moving the robot camera to see the object from another angle.
Further, since 3D meshes are available for each object, it may be possible to extract shape-level information using a PointNet~\cite{qi2017pointnet++}, similar to experiments attempted by ShapeGlot~\cite{shapeglot}.
Because we target physical robot applications, we may be able to tie language to sets of \textit{graspable points} encoding gripper orientation and position~\cite{acronym2020} for each object.

\clearpage

\section*{Acknowledgements}
\label{sec:acks}
We would like to thank the anonymous CoRL 2021 reviewers for their feedback and suggestions.
This paper was made stronger by the review and revise model adopted by CoRL this year via OpenReview.
This work was funded in part by ONR under award \#1140209-405780.

\bibliography{main}

\clearpage
\section{Supplementary Material}

\subsection{Amazon Mechanical Turk Experiments}
\label{ssec:amt}

In total, we collected 50,594 referring expression annotations from Amazon Mechanical Turk workers.
We ran annotation tasks, qualification tasks, and validation tasks in parallel, gathering data iteratively.
In total, the annotations cost around \$10,800 USD.

\paragraph{Annotation Interface}

\begin{figure}[t]
    \centering
    \includegraphics[width=1\textwidth]{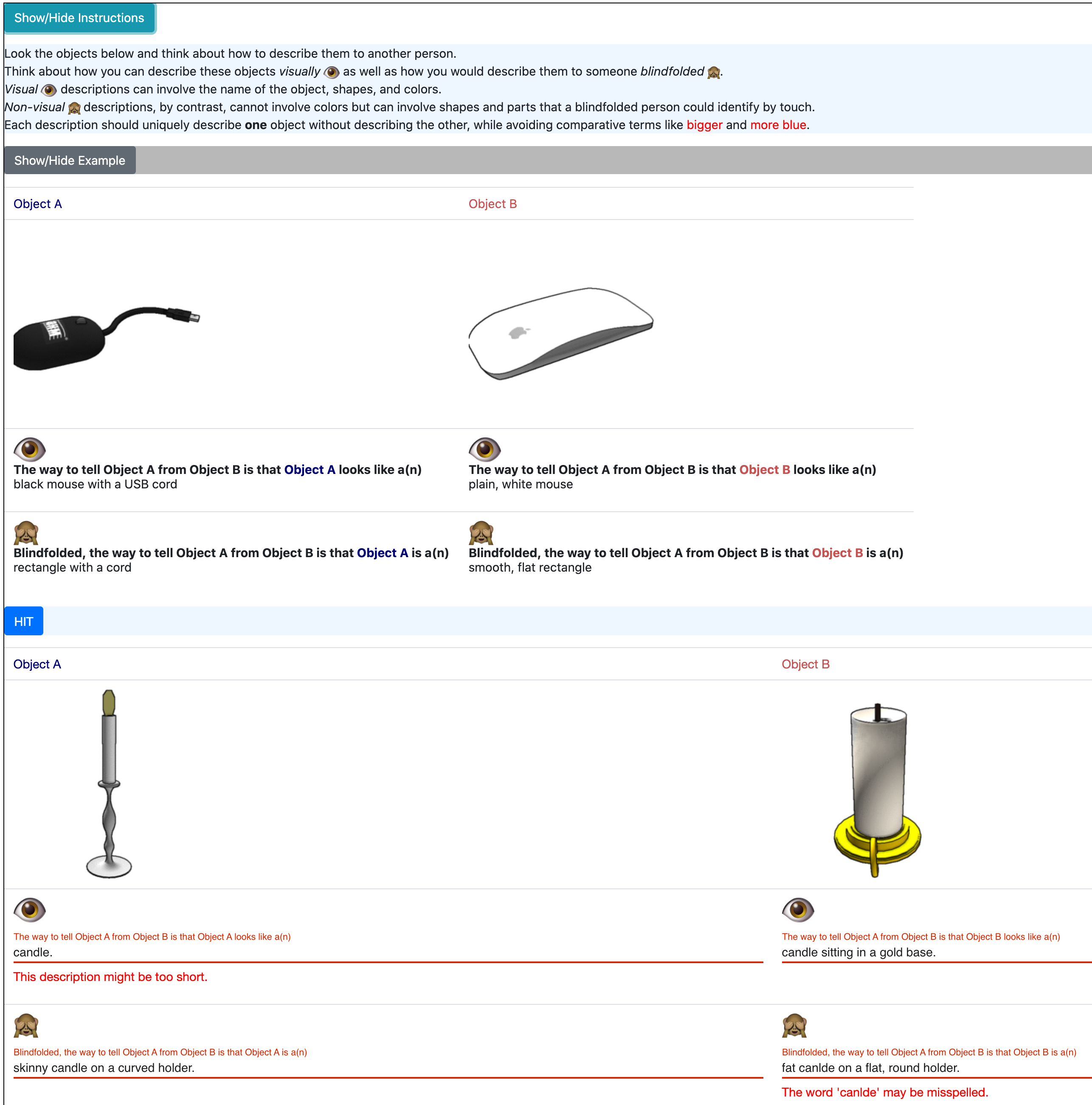}
    \caption{Workers were given instructions about the task as well as an example of how to complete it, then asked to fill in the blanks for \vis\ and \nonvis\ descriptions of two objects.
    In this interface example, warnings are given for a misspelled word and a short description.}
    \label{fig:annotation_interface}
\end{figure}

Workers wrote referring expressions conditioned on a pair of objects per HIT (Human Intelligence Task).
The interface for this annotation task is shown in Figure~\ref{fig:annotation_interface}.
Two expressions described one object, and two described the other.
For each object, one expression was primed to be \vis\ in nature, while the other was primed to be \nonvis.

During annotation, warnings were given for misspelled words, descriptions being too short, and for having substantial overlap with one another.
Workers were not prevented from submitting HITs with warnings, but workers who accrued many warnings were automatically disqualified from the task.

\paragraph{Quality Control}

Workers first qualified for our annotation task by completing a fixed qualification HIT.
In the qualification HIT, users read the instructions and wrote visual and non-visual descriptions for the example objects (\objclass{ComputerMice}) shown in Figure~\ref{fig:annotation_interface}.
To submit the qualification HIT, users had to produce four referring expressions that created no automatic warnings.

Each referring expression was validated by a secondary HIT to ensure the \dataset\ task has a high human success rate.
For every referring expression, two votes were gotten from qualified workers asked to choose which of the two objects was the referent.
Object positions (left versus right) were randomly scrambled to identify and remove referring expressions that made use of the object position on the screen at annotation time.
Validation workers were asked to select the object that the referring expression best described, and had the option to select that it described both equally or described neither.
Votes were marked as endorsements for the referring expression only if the correct object was selected.
If the two votes disagreed, a tie-breaker vote was collected.
Referring expressions with a majority of endorsing votes were kept.
In total, 5,018 annotations were flagged as unreliable, representing about 9\% of the total annotations gathered.
New annotation HITs were launched to replace such unreliable expressions.

Workers exhibiting poor performance, estimated through automated metrics, were disqualified from the task, and could not be re-qualified by taking the qualification task again.
Workers were also automatically disqualified for the annotation task if the average number of problematic expressions (those given on-screen warnings, as exhibited in Figure~\ref{fig:annotation_interface}) or number of rejected expressions (as determined by validation workers) became an outlier.
Workers were automatically disqualified for the validation task if the average number of times they were the minority voter in a disagreement became an outlier.
Outliers were identified as those whose averages on these metrics were more than one standard deviation above the mean, or 2 standard deviations in the case of automated warnings.

\subsection{\dataset\ Analysis}

\paragraph{Additional Examples}

\begin{figure}[t]
\centering
\begin{scriptsize}
\begin{tabular}{ll@{\hspace{4em}}ll}
    \multicolumn{2}{c}{\raisebox{-.5\height}{\includegraphics[width=0.2\textwidth]{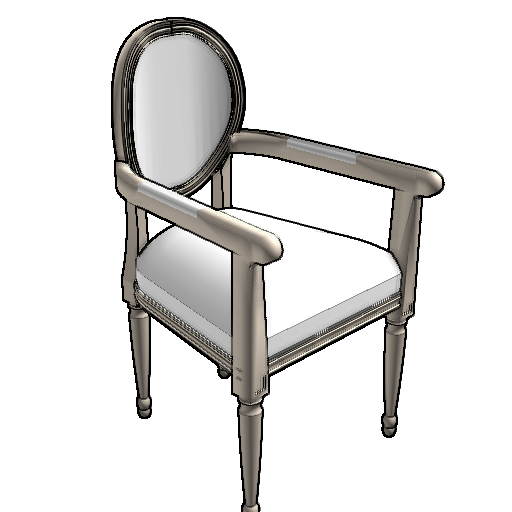}}} & \multicolumn{2}{c}{\raisebox{-.5\height}{\includegraphics[width=0.2\textwidth]{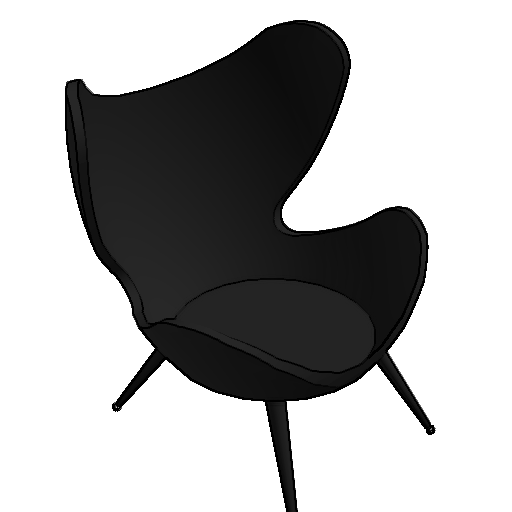}}} \\
    \textbf{\Vis} & \textbf{\Nonvis} & \textbf{\Vis} & \textbf{\Nonvis} \\
    classic armchair with & oval back, and & \multirow{2}{*}{mod black chair} & scoop shaped like puzzle \\
    \phantom{  }white seat & \phantom{  }vertical legs & & \phantom{  }piece, on four angled legs \\
    \multirow{2}{*}{white armchair} & seat cushion circular back & \multirow{2}{*}{black chair} & lowset chair with 4 \\
    & \phantom{  }rest and 4 carved wood legs & & \phantom{  }angled out legs \\
    white chair & has straight legs & black chair & has legs that flare out \\
    \midrule
    \multicolumn{2}{c}{\raisebox{-.5\height}{\includegraphics[width=0.2\textwidth]{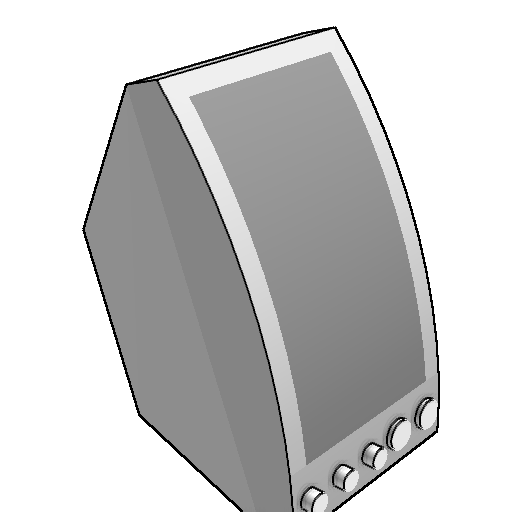}}} & \multicolumn{2}{c}{\raisebox{-.5\height}{\includegraphics[width=0.2\textwidth]{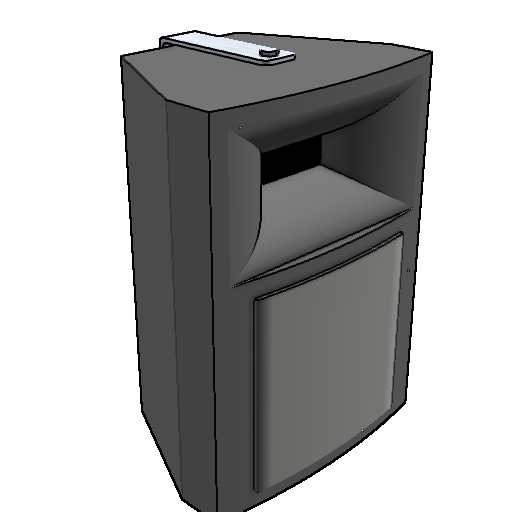}}} \\
    \textbf{\Vis} & \textbf{\Nonvis} & \textbf{\Vis} & \textbf{\Nonvis} \\
    \multirow{2}{*}{speaker with five knobs} & \multirow{2}{*}{circles on the face} & \multirow{2}{*}{rectangular speaker} & rectangle that becomes \\
    & & & \phantom{  }more narrow \\
    light grey speaker with & five small circles projected & dark grey speaker metal & square with square hole towards \\
    \phantom{  }knobs on front & \phantom{  }off front & \phantom{  }stripe down back & \phantom{  }top and bolt on top \\
    \multirow{2}{*}{light gray speaker} & curved speaker with five & \multirow{2}{*}{tall, dark grey speaker} & tall trapezoid shaped speaker \\
    & \phantom{  }knobs on the front & & \phantom{  }with hole near the top \\
    \\
\end{tabular}
\end{scriptsize}
\caption{
    Samples of object pairs and the object referring expressions in \dataset.
}
\label{fig:object_and_re_samples_2}
\end{figure}

\begin{figure}[t]
\centering
\begin{scriptsize}
\begin{tabular}{ll@{\hspace{4em}}ll}
    \multicolumn{2}{c}{\raisebox{-.5\height}{\includegraphics[width=0.2\textwidth]{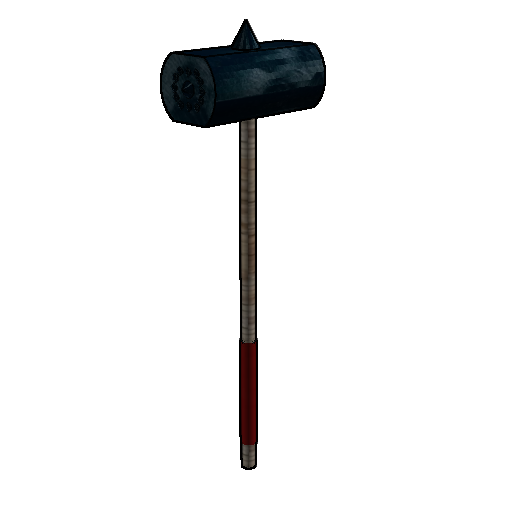}}} & \multicolumn{2}{c}{\raisebox{-.5\height}{\includegraphics[width=0.2\textwidth]{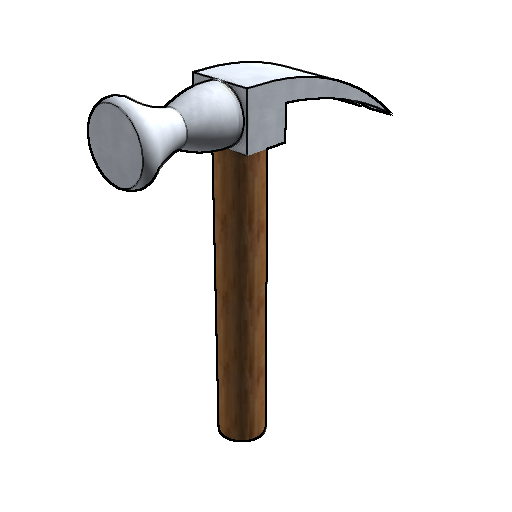}}} \\
    \textbf{\Vis} & \textbf{\Nonvis} & \textbf{\Vis} & \textbf{\Nonvis} \\
    \multirow{2}{*}{black tool.} & sharp top and sharp & hammer with wooden & long cylinder with two \\
    & \phantom{  }end of tool. & \phantom{  }handle & \phantom{  }curved wedges \\
    black mallet & thin handle & hammer silver & claw on hammer. \\
    \multirow{2}{*}{black mallet} & thin rod with pointy & \multirow{2}{*}{claw hammer} & \multirow{2}{*}{wide brown handle} \\
    & \phantom{  }cylinder on top & & \\
    \midrule
    \multicolumn{2}{c}{\raisebox{-.5\height}{\includegraphics[width=0.2\textwidth]{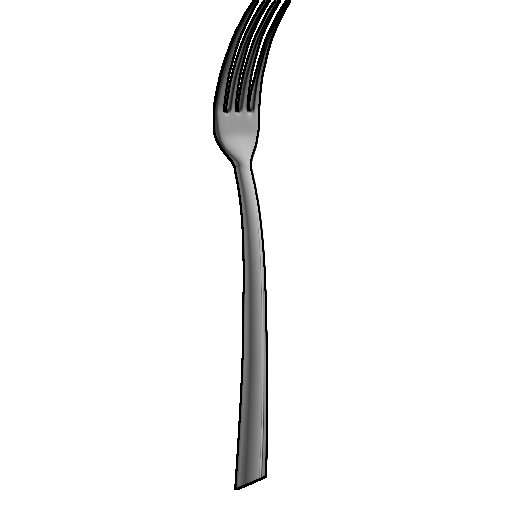}}} & \multicolumn{2}{c}{\raisebox{-.5\height}{\includegraphics[width=0.2\textwidth]{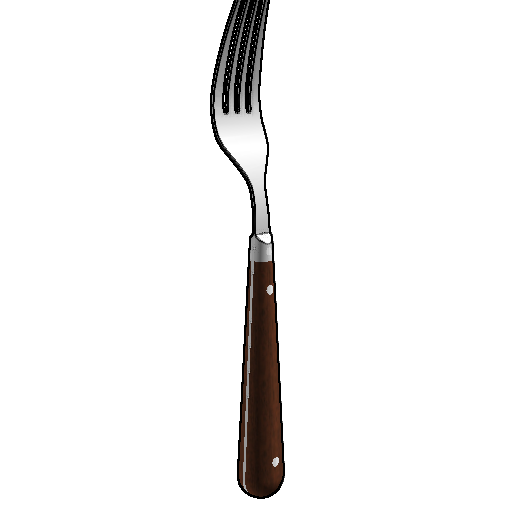}}} \\
    \textbf{\Vis} & \textbf{\Nonvis} & \textbf{\Vis} & \textbf{\Nonvis} \\
    metal fork & square bottom & wooden handle & rounded bottom \\
    single piece fork & all metal fork & metal fork with brown handle & wood handle on fork \\
    \multirow{2}{*}{silver fork} & \multirow{2}{*}{Handle is smooth} & \multirow{2}{*}{fork with brown handle.} & handle is larger than connection \\
    & & & \phantom{  }at neck of fork. \\
    \\
\end{tabular}
\end{scriptsize}
\caption{
    Samples of object pairs and the object referring expressions in \dataset.
}
\label{fig:object_and_re_samples_3}
\end{figure}

Figures~\ref{fig:object_and_re_samples_2} and~\ref{fig:object_and_re_samples_3} show additional examples of object pairs shown to workers and the resulting referring expressions gathered for \dataset.

\paragraph{Referring Expression Token Diversity}

\begin{figure}[t]
    \centering
    \includegraphics[width=1\textwidth]{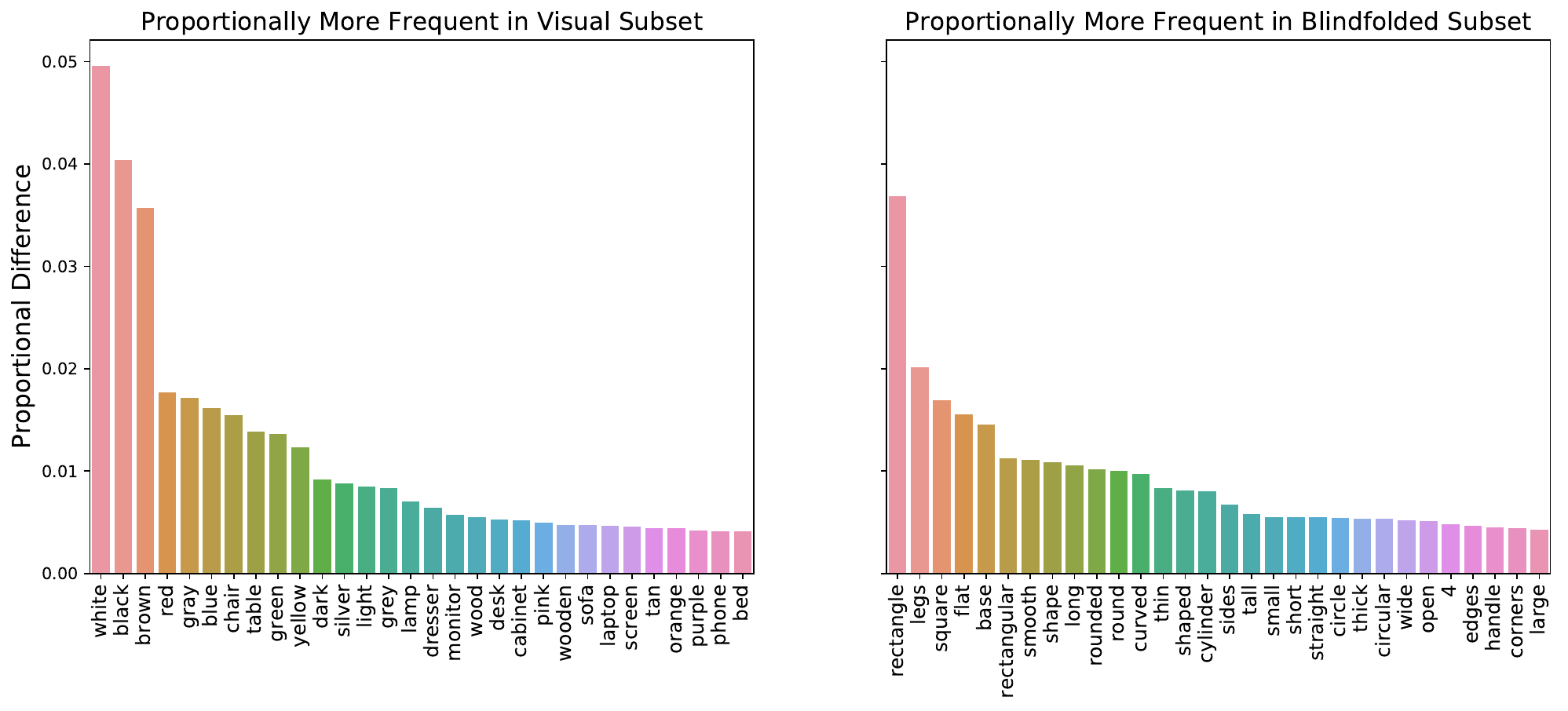}
    \caption{The top 30 token types (omitting stop words) with the highest proportional frequency difference between \vis\ and \nonvis\ primed referring expressions.
    Color and category words like \textquoteinline{white}, \textquoteinline{red}, and \textquoteinline{table} appear more frequently in \vis\ descriptions (Left), in contrast to shape and part words like \textquoteinline{rectangle} and \textquoteinline{legs} in \nonvis\ descriptions (Right).}
    \label{fig:re_prop_diffs}
\end{figure}

There are 216,146 unique token types across all referring expressions, with an average of $4.27\pm 2.07$ tokens per expression.
However, \vis\ expressions have less token diversity (93,733 types) and are shorter ($3.63\pm 1.72$ tokens) than \nonvis\ expressions (122,413 types, $4.95\pm 2.19$).
\Vis\ descriptions largely focus on color and object category, while \nonvis\ expressions include shape and part descriptions (Figure~\ref{fig:re_prop_diffs}).

\subsection{Data Folds}
\label{ssec:data_folds}

We split the data into train, validation, and test folds by ShapeNet category.
We ensure that closely related categories such as \objclass{2Shelves} and \objclass{3Shelves} or \objclass{DiningTable} and \objclass{AccentTable} are contained to a single fold.
To do so, we mapped each category to a single descriptive word, such as ``shelves,'' then iteratively assigned each category to the fold with which it had maximum cosine similarity in GloVe~\cite{glove} space according to that word.
For example, all shelf-related and bed-related categories were sorted into the train and test folds, respectively.

Objects are annotated in pairs, and these pairs are primarily drawn from the same ShapeNet category.
However, due to single-instance categories and the iterative data collection process, there are data pairs of objects paired with others outside of their category.
These \textit{cross-category} object pairs comprise only about 4\% of the 6,019 pairs used to gather object referring expressions.
Cross-category object pairs were assigned to a fold only if both categories were assigned to that fold.

To widen the available training data for the test fold, models can be trained on both the training and validation fold before evaluating on the held-out test fold.
When training to evaluate on the test fold, cross-category object pairs where one category is in train and one is in validation can also be used (only 13 object pairs fit this description).
Cross-category object pairs where one category was in the test fold and the other was not cannot be used for training, validation, or testing (only 59 object pairs fit this description).
Table~\ref{tab:full_fold_stats} presents a full summary of folds considering these cross-category pairs.

\begin{table}[t]
\centering
\begin{tabular}{lrrrr}
    \textbf{Pair Folds} & \textbf{\# Categories} & \textbf{\# Objects} & \textbf{\# Referring Expressions} & \textbf{\% Dataset Size} \\
    \toprule
    Train-Train & 207 & 6153 & 39104 & 78.0\phantom{0} \\
    Val-Val & 7 & 371 & 2304 & \phantom{0}4.6\phantom{0} \\
    Train-Val & - & 25 & 76 & \phantom{0}0.15 \\
    Test-Test & 48 & 1357 & 8751 & 17.4\phantom{0} \\
    \bottomrule \\
\end{tabular}
\caption{
    Fold summaries in the \dataset\ benchmark by pairs of object categories.
    The uniform folds, like ``Train-Train'', correspond to the folds of Table~\ref{tab:fold_stats}.
    The ``Train-Val'' fold can be used when training a final model for evaluation on the ``Test-Test'' fold, because those models can be trained on both training and validation data.
    ``Train-Val'' pairs are all cross-category, hence the `-' entry for its number of category pairs.
}
\label{tab:full_fold_stats}
\end{table}

\subsection{\vilbert\ Model}
\label{ssec:vilbert}

While our \scorer\ module uses a CLIP~\cite{clip} backbone, we also experimented with a \vilbert~\cite{lu2019vilbert} backbone.
\vilbert\ training and inference are slower, but 3D objects could be encoded via a single tiled image because \vilbert\ attends to individual bounding boxes in input images.
We compare CLIP and \vilbert\ performance on full-object (all view) encodings and find that they perform similarly, choosing CLIP as the backbone for \scorer\ for training speed.

\begin{figure}[t]
    \centering
    \fbox{\includegraphics[width=0.35\textwidth]{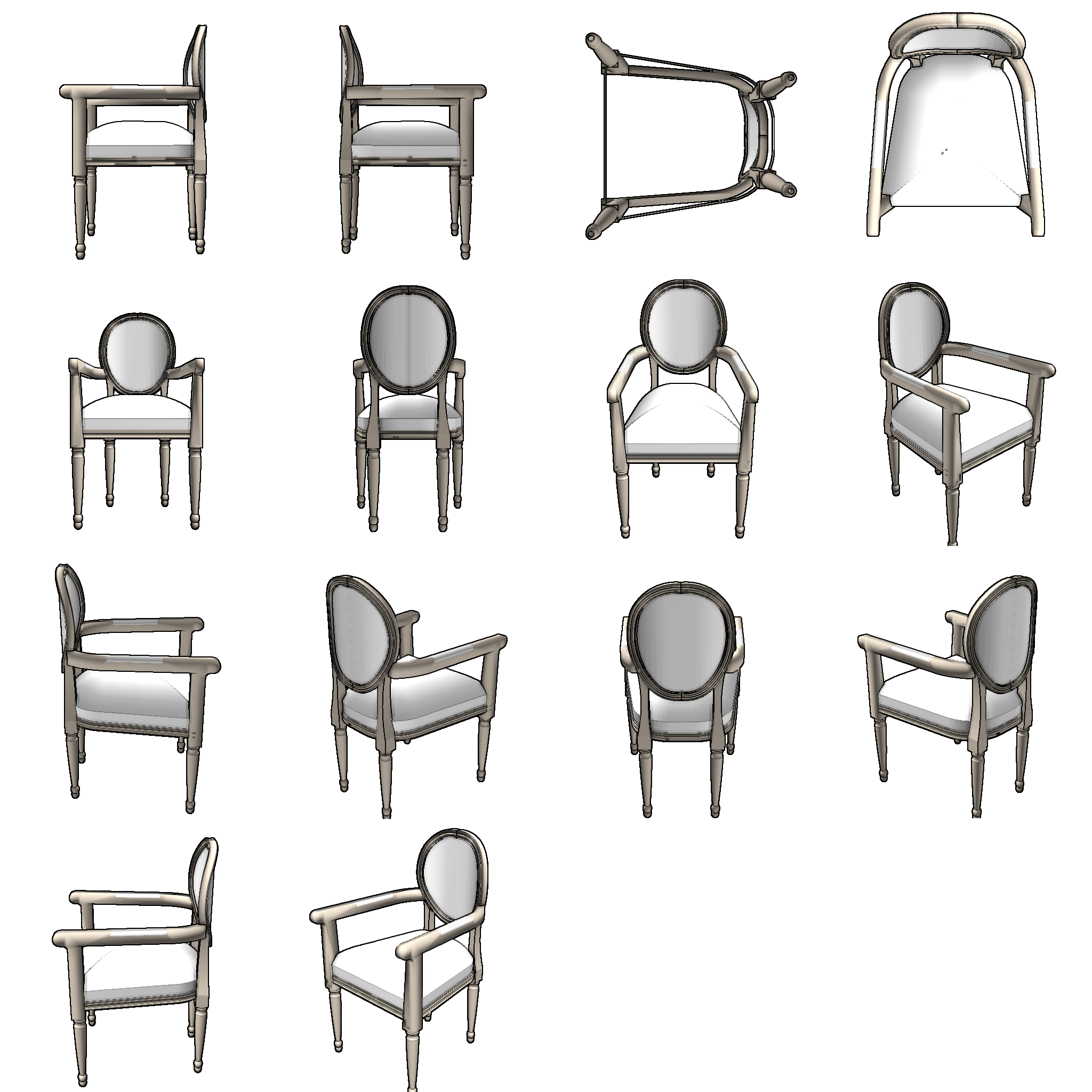}}
    \hspace{20pt}
    \fbox{\includegraphics[width=0.35\textwidth]{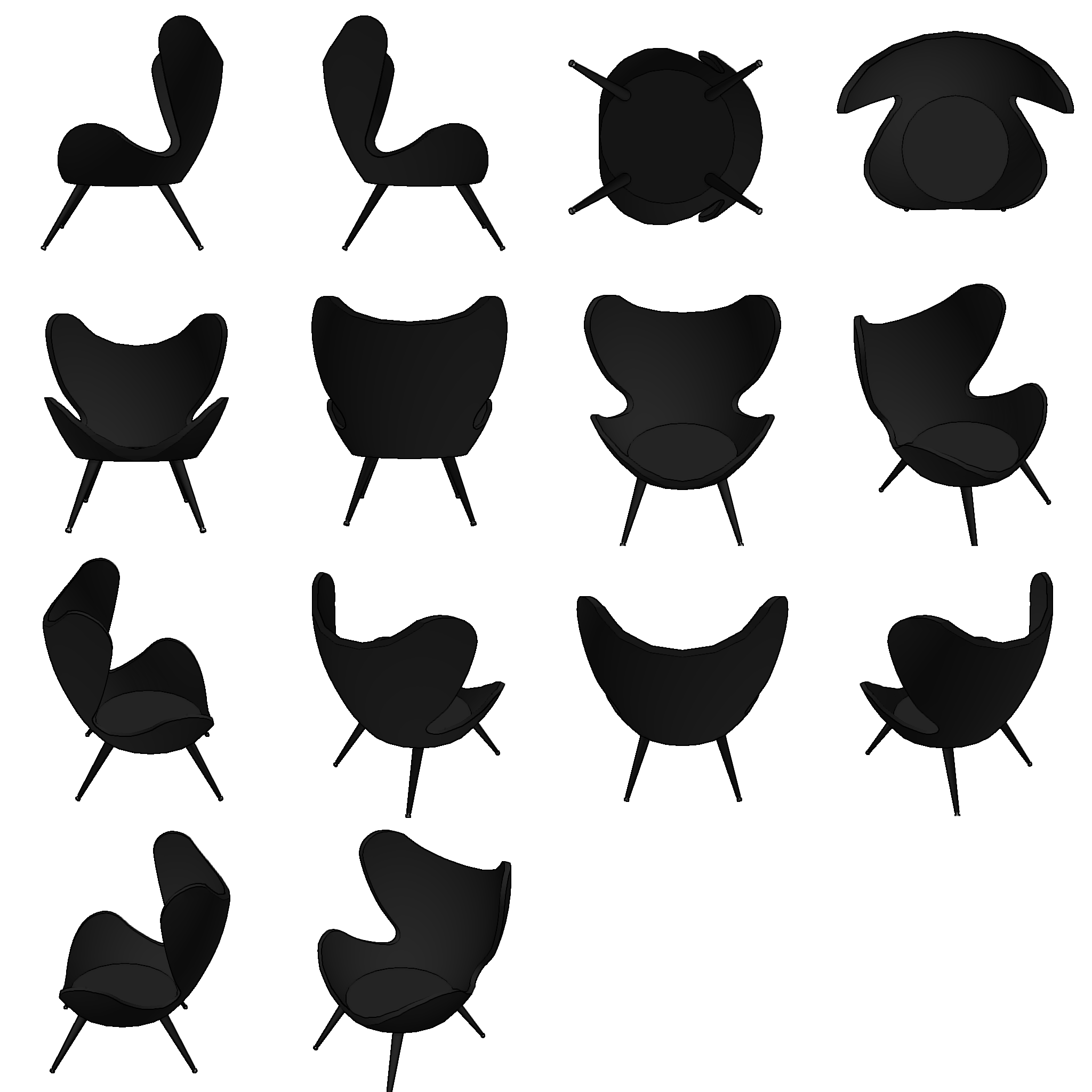}}
    \caption{\vilbert\ takes in an image and a language expression.
    To represent a 3D model, we tile 14 views of the object into a single image.
    An instance of the \dataset\ task is then represented by a language expression, such as \textquoteinline{lowset chair with 4 angled out legs}, paired with two such tiled images.
    The model must predict the image that the expression refers to.}
    \label{fig:obj_for_vilbert}
\end{figure}

\paragraph{\vilbert.}
We encoded each object as a set of 14 views from different angles as provided by ShapeNetSem (Figure~\ref{fig:obj_for_vilbert}).
We provide \vilbert\ the gold-standard bounding boxes for these 14 views during feature extraction, removing potential pipeline errors from the pretrained object detector~\cite{Lu_2020_CVPR}.
Each position in the tiled image represents a fixed camera angle from which the object is rendered, so spatial information encoded by the bounding boxes is consistent across objects; for example, the model may learn to look at the fourth image in the top row to assess facts about the ``top'' of the object.
In Table~\ref{tab:vilbert_results}, we compare \vilbert\ models that are not pretrained on any tasks versus those pretrained only on conceptual captions~\cite{sharma2018conceptual} or on 12-in-1 task pretraining~\cite{Lu_2020_CVPR}.
We find that the base \vilbert\ architecture without task pretraining is unable to learn anything meaningful with only our task data, and so only run this evaluation on the validation fold.
Note that this model still includes a pretrained BERT~\cite{devlin:naacl19} and ResNet~\cite{resnet} backbone, but without task-oriented training is too parameter-heavy to learn on our data.
\vilbert\ pretrained on conceptual captions achieves nearly as high performance as the model pretrained on 12 language and vision tasks, indicating that this large amount of aligned language and vision data is sufficient to initialize better parameters for our task.

\begin{table}[t]
\centering
\begin{tabular}{cll>{\raggedleft\arraybackslash}p{1cm}>{\raggedleft\arraybackslash}p{1cm}>{\raggedleft\arraybackslash}p{1cm}}
    & & & \multicolumn{3}{c}{\phantom{000}\textbf{Accuracy (\%) $\uparrow$}} \\
    \textbf{Fold} & \textbf{Model} & \textbf{Task Pretraining} & \textbf{All} & \textbf{Vis$\subset$} & \textbf{Blind$\subset$} \\
\toprule
    \multirow{3}{*}{\rotatebox[origin=c]{90}{Val}} & \vilbert & None; Architecture Only & $49.0$ & $48.8$ & $49.1$ \\
    & \vilbert & Conceptual Captions & $82.6$ & $89.0$ & $76.1$ \\
    & \vilbert & 12-in-1 & $\pmb{83.1}$ & $\pmb{89.5}$ & $\pmb{76.6}$ \\
    \midrule
    \multirow{2}{*}{\rotatebox[origin=c]{90}{Test}} & \vilbert & Conceptual Captions & $75.1$ & $79.4$ & $70.5$ \\
    & \vilbert & 12-in-1 & $\pmb{76.6}$ & $\pmb{80.2}$ & $\pmb{73.0}$ \\
    \bottomrule \\
\end{tabular}
\caption{
    \vilbert\ model accuracy on the \dataset\ benchmark under different pretraining regimes.
}
\label{tab:vilbert_results}
\end{table}

\subsection{Multi-view Pooling Operations}
\label{ssec:pooling}

\begin{table}[t]
   \centering
    \begin{tabular}{lclrrr}
                    & & & \multicolumn{3}{c}{\textbf{Validation}} \\
                    Model & Views & Pooling & \multicolumn{1}{c}{\Vis$\subset$} & \multicolumn{1}{c}{\NV$\subset$} & \multicolumn{1}{c}{All} \\
    \toprule
    CLIP & All & \texttt{maxpool} & 83.7 (0.0) & 65.2 (0.0) & 74.5 (0.0) \\
    CLIP & All & \texttt{meanpool} & \textbf{85.1} (0.0) & \textbf{65.7} (0.0) & \textbf{75.5} (0.0) \\
    \midrule
    \scorer & All & \texttt{maxpool} & 89.2 (0.9) & 75.2 (0.7) & 82.2 (0.4) \\
    \scorer & All & \texttt{meanpool} & \textbf{90.1} (0.7) & \textbf{75.6} (0.7) & \textbf{82.9} (0.4) \\
    \bottomrule \\
    \end{tabular}
    \caption{We experiment with meanpool as an alternative to maxpool for considering multiple object views at once via CLIP zero shot scoring and fine-tuned \scorer\ approaches.
    }
    \label{tab:snare_pooling_results}
\end{table}

In Table~\ref{tab:snare_pooling_results}, we compare \texttt{meanpool} and \texttt{maxpool} operations to unite the vector representations of the 8 views covering each object.
Using a Welch's two-tailed $t$-test, we compared the average accuracy across 10 seeds of training of \scorer\ trained with these pooling operations, and found that \texttt{meanpool} yields statistically significantly higher performance than \texttt{maxpool}.\footnote{$p=0.0022 < 0.01$; Bonferroni multiple-comparison correction applied to threshold $0.05$ because we run five such tests in total.}
Our experiments currently use \texttt{maxpool} to combine two for both \scorer\ and \model, and so this ablation indicates a further performance boost may be possible by re-training models with \texttt{meanpool}.

\subsection{Additional Robot Demonstrations}
\label{ssec:additional_robot}

\begin{figure}[t]
    \centering
    \includegraphics[width=.8\textwidth]{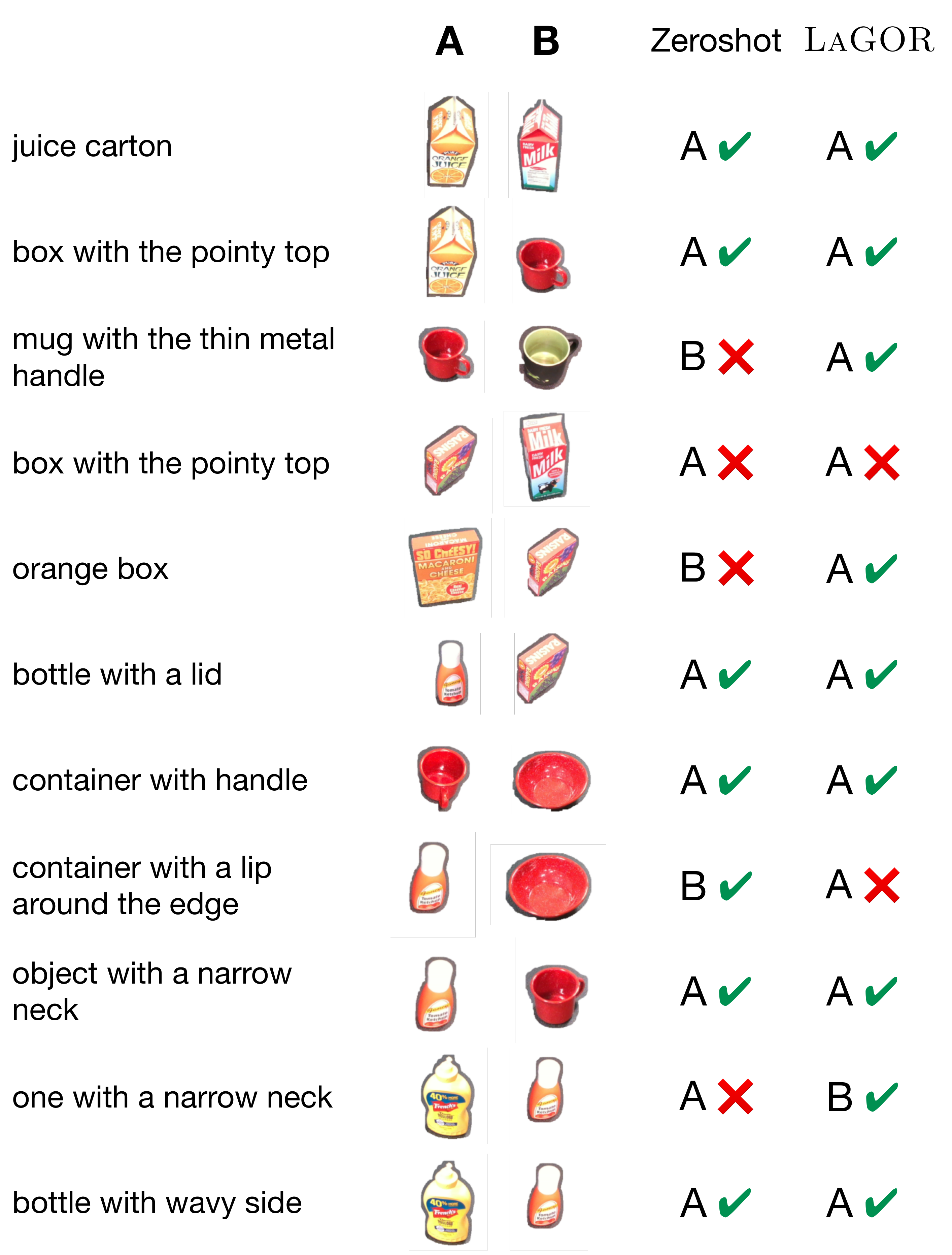}
    \caption{Examples of zeroshot CLIP versus \model\ decisions on which object a referring expression references based on segmented detections.
    Zeroshot CLIP uses only one view of each candidate objects, while \model\ uses two views from different angles.}
    \label{fig:additional_robot}
\end{figure}

Figure~\ref{fig:additional_robot} gives 11 referring expressions against pairs of objects and the zeroshot CLIP versus trained \model\ object picks.
The robot recorded two ``clean'' images for each object, with the background removed, to make the input images similar to those seen at training time.
In order to get clean masks, we performed a dilation operation on the masks, resulting in a thin border of background around each one.
By adding segmentation, this pipeline should allow us to select individual objects from a cluttered scene. 
As such, it represents closely how \model\ could be used on a real robotic system for household pick and place tasks.

\end{document}